\relax
\documentclass[letterpaper]{article} 
\usepackage{aaai20}  
\usepackage{times}  
\usepackage{helvet}  
\usepackage{courier}  
\usepackage{url}  
\usepackage{graphicx}  
\usepackage{amsfonts}
\usepackage{courier}  
\usepackage{url}  
\usepackage{graphicx}  
\usepackage{amsfonts}
\usepackage{amssymb}
\usepackage{xspace}
\usepackage{booktabs}
\usepackage{amsmath,bm}
\usepackage{dsfont}
\usepackage{multirow}
\usepackage[autostyle, english = american]{csquotes}
\usepackage[ruled,linesnumbered]{algorithm2e}
\usepackage{tabularx}
\usepackage{enumerate}
\usepackage{enumitem}
\usepackage{color}
\usepackage{xcolor}
\usepackage{xspace}
\usepackage[noend]{algpseudocode}
\usepackage{multirow}
\usepackage[flushleft]{threeparttable}

\MakeOuterQuote{"}
\usepackage{ntheorem,lipsum} 
\theorembodyfont{\upshape} 

\newcommand{\mname}{\texttt{CLARA}\xspace}
\begin{document}
%


\title{\mname: Clinical Report Auto-completion}

\author{
\Large \textbf{Siddharth Biswal\textsuperscript{\rm 1,2}, Cao Xiao\textsuperscript{\rm 1}, Lucas M. Glass\textsuperscript{\rm 1},  M. Brandon Westover \textsuperscript{\rm 3}, and Jimeng Sun\textsuperscript{\rm 2}}\\
\textsuperscript{\rm 1} Analytic Center of Excellence, IQVIA, Cambridge, USA \\
\textsuperscript{\rm 2}  Computational Science and Engineering, Georgia Institute of Technology, Atlanta, USA 
\textsuperscript{\rm 3}  Massachusetts General Hospital, Boston, MA
}

\maketitle
\begin{abstract} 
Generating clinical reports from raw recordings such as X-rays and electroencephalogram (EEG) is an essential and routine task for doctors. However, it is often time-consuming to write accurate and detailed reports.
Most existing methods try to generate the whole reports from the raw input with limited success because 1) generated reports often contain errors that need manual review and correction, 2) it does not save time when doctors want to write additional information into the report, and 3) the generated reports are not customized based on individual doctors' preference. We propose {\it CL}inic{\it A}l {\it R}eport {\it A}uto-completion (\mname), an interactive method that generates reports in a sentence by sentence fashion based on doctors' anchor words and partially completed sentences. 
\mname searches for most relevant sentences from existing reports as the template for the current report. The retrieved sentences are sequentially modified by combining with the input feature representations to create the final report. In our experimental evaluation \mname achieved 0.393 CIDEr and 0.248 BLEU-4 on X-ray reports and 0.482 CIDEr and 0.491 BLEU-4 for EEG reports for sentence-level generation, which is up to 35\% improvement over the best baseline. Also via our qualitative evaluation, \mname is shown to produce reports which have a significantly higher level of approval by doctors in a user study (3.74 out of 5 for \mname vs 2.52 out of 5 for the baseline).

\end{abstract}
\newcommand{\T}[1]{\ensuremath{\mathcal{#1}}} 
\newcommand{\M}[1]{\ensuremath{\mathbf{#1}}} 
\newcommand{\V}[1]{\ensuremath{\mathbf{#1}}} 

\newcommand{\Ab}{\mathbf{A}}
\newcommand{\Bb}{\mathbf{B}}
\newcommand{\Cb}{\mathbf{C}}
\newcommand{\Db}{\mathbf{D}}
\newcommand{\Eb}{\mathbf{E}}
\newcommand{\Fb}{\mathbf{F}}
\newcommand{\Gb}{\mathbf{G}}
\newcommand{\Hb}{\mathbf{H}}
\newcommand{\Ib}{\mathbf{I}}
\newcommand{\Jb}{\mathbf{J}}
\newcommand{\Kb}{\mathbf{K}}
\newcommand{\Lb}{\mathbf{L}}
\newcommand{\Mb}{\mathbf{M}}
\newcommand{\Nb}{\mathbf{N}}
\newcommand{\Ob}{\mathbf{O}}
\newcommand{\Pb}{\mathbf{P}}
\newcommand{\Qb}{\mathbf{Q}}
\newcommand{\Rb}{\mathbf{R}}
\newcommand{\Sb}{\mathbf{S}}
\newcommand{\Tb}{\mathbf{T}}
\newcommand{\Ub}{\mathbf{U}}
\newcommand{\Vb}{\mathbf{V}}
\newcommand{\Xb}{\mathbf{X}}
\newcommand{\Yb}{\mathbf{Y}}
\newcommand{\Zb}{\mathbf{Z}}

\newcommand{\ab}{\mathbf{a}}
\newcommand{\bb}{\mathbf{b}}
\newcommand{\cb}{\mathbf{c}}
\newcommand{\db}{\mathbf{d}}
\newcommand{\eb}{\mathbf{e}}
\newcommand{\fb}{\mathbf{f}}
\newcommand{\gb}{\mathbf{g}}
\newcommand{\hb}{\mathbf{h}}
\newcommand{\ib}{\mathbf{i}}
\newcommand{\jb}{\mathbf{j}}
\newcommand{\kb}{\mathbf{k}}
\newcommand{\lb}{\mathbf{l}}
\newcommand{\mb}{\mathbf{m}}
\newcommand{\nb}{\mathbf{n}}
\newcommand{\ob}{\mathbf{o}}
\newcommand{\pb}{\mathbf{p}}
\newcommand{\qb}{\mathbf{q}}
\newcommand{\rb}{\mathbf{r}}
\newcommand{\sbb}{\mathbf{s}}
\newcommand{\tb}{\mathbf{t}}
\newcommand{\ub}{\mathbf{u}}
\newcommand{\vb}{\mathbf{v}}
\newcommand{\wb}{\mathbf{w}}
\newcommand{\xb}{\mathbf{x}}
\newcommand{\yb}{\mathbf{y}}
\newcommand{\zb}{\mathbf{z}}
\newcommand{\xseq}{x_{n,1:\text{T}_n}}
\newcommand{\eye}{\mathbf{I}}
\newcommand{\real}{\mathbb{R}}
\newcommand{\data}{\mathcal{D}}
\newcommand{\normal}{\mathcal{N}}
\newcommand{\E}{\mathbb{E}}
\newcommand{\hide}[1]{}
\newcommand{\KL}[2]{\text{KL}(#1 \text{ } || \text{ }#2)}
\newcommand{\ent}[1]{\mathbb{H}({#1})}
\newcommand{\elbo}{\mathcal{L}}
\newcommand{\squash}{\xi}
\newcommand{\grad}[1]{\nabla_{#1}}
\newcommand{\es}{s}
\newcommand{\rpm}{\raisebox{.2ex}{$\scriptstyle\pm$}}


\newcommand{\reminder}[1]{{\textsf{\textcolor{red}{[#1]}}}}

\section{Introduction}

Medical imaging or neural recordings (e.g., X-ray images or EEG) are widely used in clinical practice for diagnosis and treatment. Typically clinical experts will visually inspect the images and signals, and then identify key disease phenotypes and compose text reports to narrate the abnormal patterns and detailed explanation of those findings.
Currently, clinical report writing is cumbersome and labor-intensive. Moreover, it requires thorough knowledge and extensive experience in understanding the image or signal patterns and their correlations with target diseases~\cite{world2004neurology}. In the age of telemedicine, more diagnostic practices can be done on the web which requires a more efficient diagnostic process. Improving the quality and efficiency of medical report writing can have a direct impact on telemedicine and healthcare on the web.

To alleviate the limitation of manual report writing, several medical image reporting generation methods~\cite{jing2017automatic} have been proposed. 
However, none of the existing works simultaneously provide the following desired properties for medical report generation.
\begin{enumerate}
    \item \textit{Align with disease phenotypes.} Medical reports describe clinical findings and diagnosis from medical images or neural recordings, which need to align with disease phenotypes and ensure the correctness of medical terminology usage.
    \item \textit{Adaptive report generation.} The generated reports need to be adapted to the preference of end-users (e.g., clinicians) for improved adoption. 
\end{enumerate}


\begin{figure}[bt]
\centering
    \includegraphics[width=1\columnwidth]{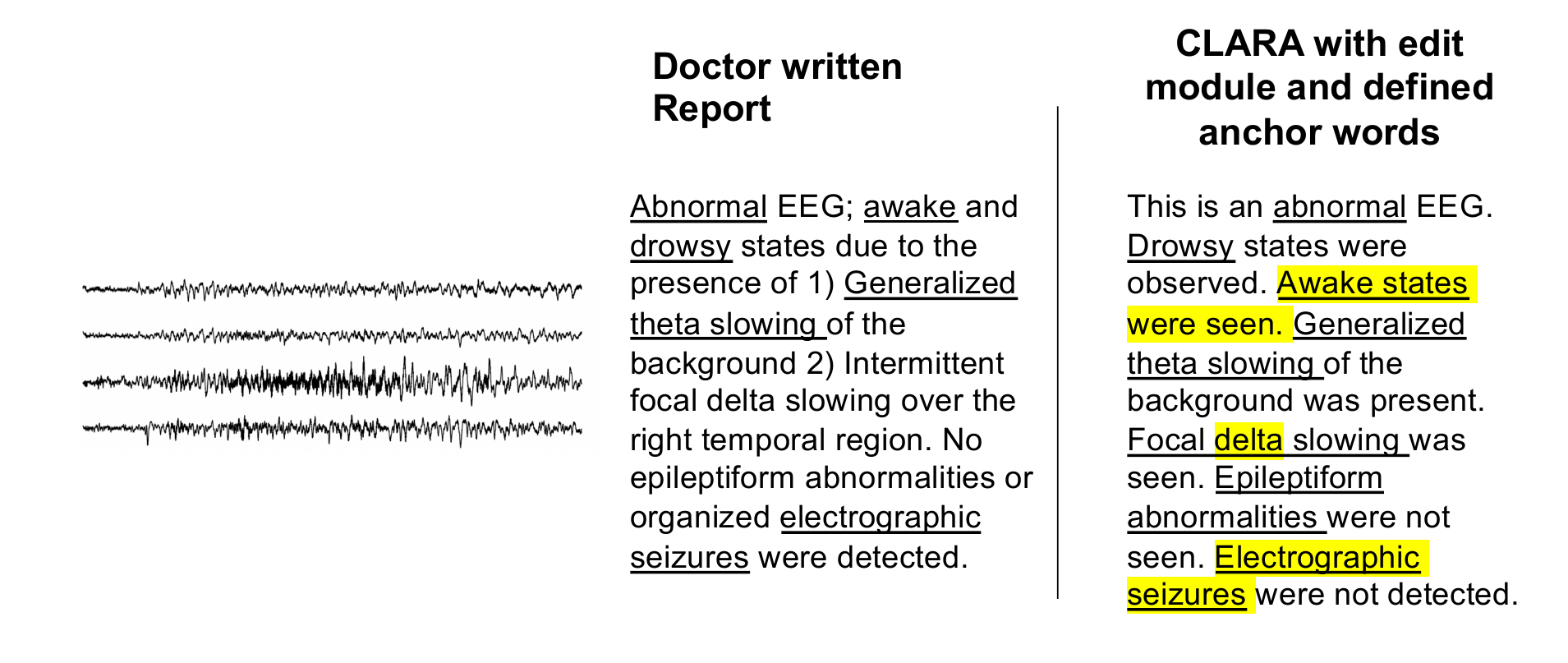}
    \vskip -1em
    \caption{\mname uses input data such as EEG or X-ray with anchor words(disease phenotypes) to produce a report sentence by sentence. In the inference mode, doctors will be able to provide the anchor words or predicted anchor words can be used to generate the report providing more control over the final output report. \mname uses different anchor words for the final report generation. In the above report {\it Abnormal EEG}, {\it Generalized slowing}, {\it Seizures}, {\it Epileptiform discharges} were used as the anchor words to generate the final report.}
    \vskip -1em
    \label{fig:userstudy}
\end{figure}
To fill the gap, we propose an interactive method named \mname to fill in the medical reports in a sentence by sentence fashion based on anchor words (disease phenotypes) and partially completed sentences (prefix text) provided by doctors. \mname adopts an adaptive retrieve-and-edit framework to progressively complete report writing with doctors' guidance. \mname constructs a prototype sentence database from all previous reports. In particular, \mname extracts the most relevant sentence templates based on user queries and then edit those sentences with the feature representation extracted from the data. In particular, the retrieval step uses an information retrieval system such as Lucene to enable fast, flexible and accurate search~\cite{bworld}. Then the edit step uses a modified version of the seq2seq method  \cite{Sutskever_Vinyals_Le_2014} to generate sentences for the current report. The latent representation of the previous sentences is adaptively used as context to generate the next sentence. In summary, \mname has the following contributions compared with other medical report generation approaches.

\begin{enumerate}
    \item \textit{Phenotype oriented.} Since \mname generated report is created using the anchor words of relevant disease phenotypes, it ensures that the report is clinically accurate. We also evaluate our method on clinical accuracy via disease phenotype classification.
    \item \textit{Interactive report generation.} Users (e.g., doctors) have more control over the generated reports via interactive guidance on a sentence by sentence level.
\end{enumerate}

We evaluate \mname on two types of clinical report writing tasks: (1) X-ray report generation that takes fixed length imaging data as input, and (2) EEG report generation that considers varying-length EEG time series as input. 
For EEG data, we evaluated our model using two datasets to test the generalizability of \mname. We show that with our \mname framework, we can achieve 0.393 CIDEr and 0.248 BLEU-4 on X-ray reports and 0.482 CIDEr and 0.491 BLEU-4 for EEG reports for sentence-level generation, which is up to 35\% improvement over the best baseline.  Compared to other methods, our \mname approach can generate more clinically meaningful reports. We show via a user study, \mname can produce more clinically acceptable reports measured through quality score metric 3.74 out of 5 for \mname vs. 2.52 out of 5 for the best baseline.
\section{Related Work}
\noindent\textbf{Image captioning} generates short descriptions of image input. There have been few attempts at solving image captioning task before the deep learning era \cite{yao2010i2t,ordonez2011im2text}. Several deep learning models were proposed for this task~\cite{vinyals2015show,karpathy2015deep}.  Many of these different image captioning frameworks proposed can be categorized into template-based, retrieval-based and novel caption generation\cite{farhadi2010every,you2016image,li2011composing,mao2014deep,lu2018neural,dai2018compositional,VenugopalanHRMD16,RennieMMRG16,chenShowFool,xu2015show}. However, they do not perform very well in generating longer paragraphs. There is limited research for generating longer captions, notably hierarchical RNN~\cite{krause2017hierarchical}. 

\noindent\textbf{Medical report generation} adapts similar ideas from image captioning to generate full medical text report based on X-ray images~\cite{jing2017automatic,li2018hybrid,ZhangXXMY17,LiuCliniaclly2019,Zhang2018,gale2018radiology,li2019knowledge}.  To improve report accuracy, researchers have utilized curated report templates to simplify the generation task~\cite{li2019knowledge,HanWLC018}. However, the generated full reports often contain errors that require significant time to correct. \mname focuses on an interactive report generation that follows the natural workflow of clinicians and led to more accurate results. \mname does not require any predefined templates but instead retrieves and adapts existing reports to generate the new one interactively. More recently, \cite{EEG2text} develops a template-based approach to generate EEG reports using a hybrid model of CNN and LSTM.

\noindent\textbf{Query auto-completion} is about expanding prefix text with related text to generate
more informative search queries. This is a well-established topic~\cite{Cai2016-mq}. Tradition query auto-completion suggests the more popular and relevant queries to the prefix text~\cite{Bar-Yossef2011-uk}. Recently neural networks models have been  used for query auto-completion task that can potentially generate new and unseen queries
using LSTM~\cite{jaech2018personalized} and hierarchical encoder-decoder~\cite{sordoni2015hierarchical}. \mname differs in terms of input for the model as our models accept multimodal input, not just short prefix text. 


\section{Method}

\subsection{Task Formulation}

\begin{figure*}[!htb]
\centering
    \includegraphics[width=1\textwidth]{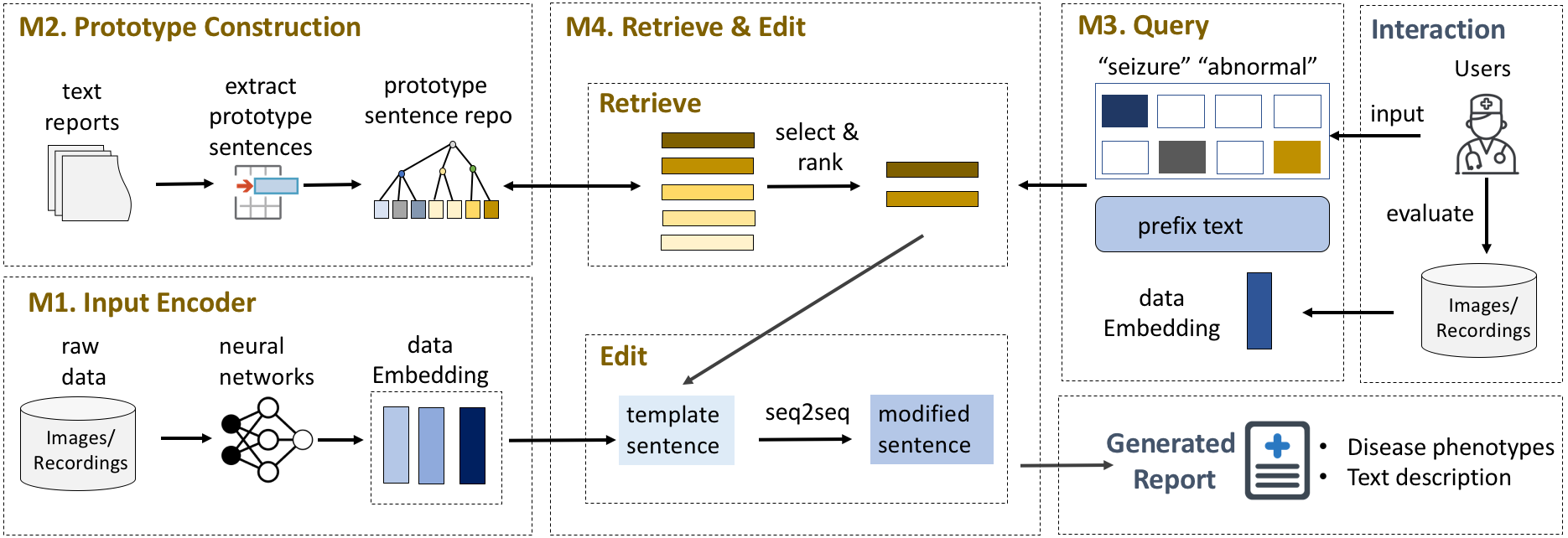}
    \caption{An overview of \mname. \mname has an input encoder module to learn embeddings from medical images or neural recordings, meanwhile, a prototype repository is constructed by indexing the unique sentences from all medical reports. Anchor words and prefix text provided by users will be served as queries to retrieve most relevant sentence templates. 
    These sentence templates will be modified by the edit module via seq2seq model to produce a new sentence for the current report. The process will repeat iteratively to generate all sentences in the report description and the associated disease phenotypes.}
    \label{fig:eeg2text_framework}
\end{figure*}

\textbf{Data}: We denote data samples as $\Db^{(i)} = \{(\Xb^{(i)},K^{(i)}, \M{P}^{(i)}, \Yb^{(i)})\}_{i=1}^I$.
In the case of EEG data, we denote  $\Xb^{(i)}\in \real^{C \times T}$ 
as the EEG record for subject $i$, where $C$ is the number of electrodes and $T$ is the number of discretized time steps per recording. 
In the case of X-ray, the input $\Xb^{(i)}\in \real^{C \times T}$ is a $C\times T$ image. $K^{(i)}$ and $\M{P}^{(i)}$ are the guidance provided by users, namely, the anchor words and prefix text for subject $i$.
These anchor words include general descriptions such as ``normal'' as well as diagnostic phenotype such as ``seizure''.
The prefix text is the first few words from each sentence in the report. \\

\noindent\textbf{Task}: In this work, we focus on generating findings (impression) section  of medical reports due to its clinical importance. Given an input sample $\Xb^{(i)}$ (X-ray or EEG), \mname generates a text report  consisting of a sequence of sentences $\Yb^{(i)}=(\Sb_{1}^{(i)},\Sb_{2}^{(i)}, \ldots, \Sb_{J}^{(i)})$ to narrate the patterns and findings in $\Xb^{(i)}$.  $\M{P}^{(i)}=(P_{1}^{(i)},P_{2}^{(i)}, \ldots, P_{J}^{(i)})$ are optional prefix texts provided by users for each sentence. Note that $\M{P}^{(i)}$ can be empty. 
\mname generates a sentence $\Sb_{j}^{(i)}$ using the data embedding of input $\M{X}^{(i)}$ and the context generated by the previous sentence $\Sb_{j-1}^{(i)}$, anchor words  $K^{(i)}$ and optional prefix text $P_j^{(i)}$. 
The notations are summarized in Table.~\ref{tab:symbol}. We have illustrated the overall \mname framework in Fig \ref{fig:eeg2text_framework}.

\begin{table}[h]
\centering
\caption{Notations used in \mname. }\label{tab:symbol}
    \begin{tabularx}{\linewidth}{ c X }
    \toprule
    \bf Notation & \bf Definition \\
     $\Db^{(i)} = \{(\Xb^{(i)},K^{(i)}, \Yb^{(i)})\}_{i=1}^I$ & $i$-th data sample,  $i=1,2... I$   \\
    \multirow{2}{*}{ $\Xb^{(i)}\in \real^{C \times T},\quad \fb^{(i)}$}  & $i$-th input sample (X-ray or EEG) and its  embedding, $i=1,2... I$
     \\
     $\Yb^{(i)}= (\Sb_{1}^{(i)},\Sb_{2}^{(i)}, \ldots, \Sb_{J}^{(i)})$ & $i$-th report,  $i=1,2... I$ 
     \\
    \multirow{2}{*}{  $\Sb_j^{(i)}$} & $j$-th sentence in the $i$-th report,  $j=1,2,\ldots, J$
    \\
    \multirow{2}{*}{ $K^{(i)}$} &  anchor words provided by  users for the $i$-th report \\
    \multirow{2}{*}{ $\M{P}^{(i)}=(P_{1}^{(i)},P_{2}^{(i)}, \ldots, P_{J}^{(i)})$} & optional prefix text  provided by users for each sentence\\
    \multirow{2}{*}{$\M{Y}_p = \{\M{Y}_{p1}, \M{Y}_{p2}, \ldots, \M{Y}_{pN}\}$} & 
    $N$ prototype sentences extracted from all reports
    \end{tabularx}
\end{table}


\subsection{The \mname Framework} 
The \mname framework comprises of the following modules.

\begin{itemize}
    \item \textbf{M1.} \textit{Input encoder module} transforms medical data such as image or EEG time series into compressed feature representations.
    \item \textbf{M2.} \textit{Prototype construction}  constructs a sentence-level repository which includes distinction sentences, their representations, writer information and frequency statistics derived from a large medical report database. This repository will be searched dynamically to provide a starting point for generating sentences in a new report. 
    \item \textbf{M3.} \textit{Query module} provides more control for the clinicians to interactively produce a customized medical report. It accepts queries from the clinicians in the form of anchor words (global context) and prefix text (local context). {\it Anchor words} are phenotype keywords associated with the entire report.  And optional {\it prefix text} are partial sentences entered by the users through interactive edit. 
    \item \textbf{M4.} \textit{ Retrieve and edit module} interactively produces report guided by users using the data representation, anchor words, and prefix text. This module sequentially performs report generation. First, the retrieve module extracts most relevant sentences from prototypes repository. Then the edit module uses a sequence-to-sequence~\cite{Sutskever_Vinyals_Le_2014} model to modify the retrieved sentences based on the data rsentation, anchor words, and prefix text.
\end{itemize}

\noindent\textbf{M1. Input Encoder Module}
This module is used to extract data embedding from the input to guide the report completion. The input can be raw measurements of X-ray or EEG. For both images $\Xb^{(i)}$ and EEG time series  $\Xb$ in the form of a sequence of EEG epochs $\Xb$ = {$\xb_1$,$\xb_{2}$,\ldots,$\xb_{T}$}, we can encode them using a convolutional neural network(CNN) to obtain image embedding $\fb^{(i)}$, or the EEG embedding $\fb^{(i)}_{t}$ for epoch $t$. 
\begin{equation}
    \fb^{(i)} = \mathbb{CNN}(\Xb^{(i)}); \quad \fb^{(i)}_{t} = \mathbb{CNN}(\xb_{i}).
\end{equation}
For X-ray imaging, the DenseNet~\cite{huang2017densely} architecture is used for CNN. For EEG, the final embedding for all epochs is the average embedding $\fb^{(i)} = \frac{1}{T}\sum_t \fb^{(i)}_{t}$. We use a CNN with convolutional-max pooling blocks for processing the EEG data into feature space. We use Rectified Linear Units(ReLUs)  activation function for these convolutional networks, and with batch normalization \cite{Szegedy2013-ym}
More detailed model configuration is provided in the experiment section. Finally, we average over these feature vectors to produce $\fb^{(i)}$ for an EEG recording associated with the sample. More sophisticated aggregations such as LSTM or attention model is considered as well but with very limited improvement. Therefore, we decide to use this simple but effective method of average embedding. The output data embedding will be fed into the retrieving step to be associated with anchor words and used to generate reports jointly. The anchor words are provided as labels.\\

\noindent\textbf{M2. Prototype Construction}
The idea here is to organize all the existing sentences from medical reports into a retrieval system as prototype sentences. We take a hybrid approach between information retrieval and deep learning to structure prototype sentences. \\

\noindent{\it Motivation:} Prototype learning~\cite{DBLP:journals/corr/SnellSZ17,DBLP:journals/corr/abs-1710-04806,li2019knowledge} and memory networks~\cite{DBLP:journals/corr/WestonCB14,sukhbaatar2015end} are different ways to incorporate data instances directly into the neural networks. The common idea is to construct a set of prototypes $\M{Y}_p = \{\M{Y}_{p1}, \M{Y}_{p2}, \ldots, \M{Y}_{pN}\}$ and their representation $\{f(p)| p \in \M{Y}_p\}$. 
Then given a new data instance $x$,  prototype learning will try to learn a representation of $x$ as $[d(f(x), f(\M{Y}_{p1})), \ldots, d(f(x),f(\M{Y}_{pN})) ]$ where $d(\cdot,\cdot)$ is a distance or similarity function. Similarly, memory network will put all those prototype representation in a memory bank and learn a similarity function between $x$ and every instance in the memory bank. 
However, there are several significant limitations to these approaches: 
1) {\bf storage and computation cost} can be large when we have a large number of prototypes. For example, we want to treat all unique sentences from a medical report database as prototypes. Every pass of the network involves a large number of distance/similarity computations.  2) {\bf static prototypes - } Often prototypes and their representations have to be fixed first before the prototype learning model can be trained. Also once the model is trained, no new prototypes can be added easily. In medical report applications, new reports are continuously being created and should be incorporated into the model without retraining from scratch. 3) {\bf computational waste - } it seems quite wasteful to conduct all the similarity computations knowing only a small fraction of prototypes are relevant for a given query. \\

\noindent{\it Approach:} We take a scalable approach to structure prototypes in \mname. We extract all sentences from a large collection of medical reports, then index these sentences to be used by a retrieval system, e.g., inverted index over the unique sentences. We also weigh those sentences based on their popularity so that frequent sentences will have higher weights to be retrieved. There are several immediate benefits of this approach: 1) we can support a large number of sentences as a typical retrieval system such as Lucene can support a web-scale corpus; 2) We are able to update the index with new documents easily so new reports can be integrated; 3) The query response is much faster than a typical prototype learning model thanks to the fast retrieval system. 
Formally, given a report corpus $\{\Yb^{(1)}, \Yb^{(2)}, \ldots, \Yb^{(I)}\}$, we map them into a set of sentence pairs $\Sb = \{(s_1, w_1), (s_2,w_2), \ldots, (s_{I'}, w_{I'})\}$ where $I$ is the number of reports and $I'$ the number of unique sentences. Then we index the set $\Sb$ with retrieval engine such as Lucene to support similarity query.\\

\noindent\textbf{M3. Query Module}
 provides interactive report auto-completion for users  to efficiently produce report sentence by sentence. It has two ways of interactions.
\begin{enumerate}[leftmargin=*]
    \item {\it Anchor words} $K$ are a set of keywords that provide a high-level context for the report. For EEG reports,  anchor words include Normal, Sleep, Seizure,  Focal Slowing, and Epileptiform. Similarly, for X-ray reports anchor words include Pneumonia, Cardiomegaly, Lung Lesion, Airspace Opacity, Edema, Pleural Effusion,  Fracture as used in~\cite{irvin2019chexpert}. 
    
    \item {\it Prefix text} $P_j^{(i)}$  specifies the partial sentence of  sentence $j$ in report $i$. This prefix text enables customization and controls from users. Note that prefix text are completely optional to \mname.
\end{enumerate}
Anchor words and prefix text are used in the Retrieve module to find relevant sentences from the prototype repository.\\

\noindent\textbf{M4. Interactive Retrieve and Edit}
 module aims to find the most relevant sentences from the prototype repository  ({\it Retrieve} phase), and then edit them to fit the current report ({\it Edit} phase). Usually, clinicians use a predefined template to draft the report in the clinical workflow. For example, the standard clinical documentation often follows a SOAP note (an acronym for subjective, objective, assessment, and plan). In this case, we seek sentence-level templates that users prefer using. Below we describe the two-phase approach that \mname uses to generate sentences for medical reports. \\

In the \textbf{retrieve} phase, we use an information retrieval system to find the most relevant sentences in the prototype repository. 
This step simulates a doctor looking up his previously written reports to identify the relevant sentences to modify. Given an anchor word $K^{(i)}_l$ and optional prefix text, this module extracts a template sentence $S_l$ from the prototype repository. Here we use the widely-adopted information retrieval system Lucene to index and search for the relevant sentences \cite{bworld,zobel2006inverted,perez2009integrating}. More details of indexing and scoring operations performed by Lucene engine are in Appendix~\ref{appendix:lucene}.
If anchor words are not available, \mname will first predict what anchor words should be there by learning a classifier from data embedding $\M{f}^{(i)}$ to anchor words $K^{(i)}$. Compared to other retrieve approach such as~\cite{li2019knowledge}, our approach is more flexible and scalable thanks to the power of retrieval systems.\\

In the \textbf{edit} phase, the retrieved sentence is modified to produce the final sentence for the current report. We adopted a sequence-to-sequence model~\cite{Sutskever_Vinyals_Le_2014} which consists of an encoder and a decoder, where the encoder projects the input to compressed representations and the decoder reconstructs the output. Here we use both the sentence template $S_l$ and the data embedding $\fb^{(i)}$ as input for the encoder and revised sentence is the output sequence. The encoder is implemented as two layer bi-directional Long short term memory network (LSTM)~\cite{hochreiter1997long}. 
The decoder is a three-layered LSTM. The decoder takes the resulting context vector $\zb_l$ as input for the generation process. Then it is concatenated with the decoder's hidden states and used to compute a softmax distribution over output words to produce the final $\yb_l$.
\begin{equation}
    \zb_l = \mathbb{LSTM_{\text{encoder}}}(S_l, \fb^{(i)},\zb_{l-1});\quad
    \yb_l = \mathbb{LSTM_{\text{decoder}}}(\zb_l) 
\end{equation}
Our \mname framework uses a sequential generation process to produce the final report. We iteratively use the previous hidden states with the encoder to enforce the context generated at each sentence to guide the next sentence generation. The anchor words and prefix texts are often included in the final report generated as these words are part of the reports. 


    
    

\section{Experiment}
We evaluate \mname framework to answer the following questions:

\begin{enumerate} [leftmargin=*]
    \item[\textbf{A}:] Can \mname generate higher quality clinical reports?
    \item[\textbf{B}:] Can the generated reports capture disease phenotypes?
    \item [\textbf{C}:] Does \mname generate better reports from clinicians'  view?
\end{enumerate}

\subsection{Experimental Setup}

\noindent\textbf{Data} We conduct experiments using the following datasets.

\noindent (1) \textit{Indiana University X-ray Data(IU X-ray)} dataset contains 7,470 images and paired reports collected. Each patient has 2 images (a frontal view and a lateral view) \cite{demner2015preparing}. The paired report contains impression, finding and indication sections. We apply some data preprocessing techniques to remove duplicates from this dataset. For X-ray reports, we only focus on findings section of the report. After extracting the findings section, we apply tokenization and keep tokens with at least 3 occurrences in the corpus resulting in 1235 tokens in total.We use the labels used by CheXpert labeler as the anchor words \cite{irvin2019chexpert}. These labels are representative of the different phenotypes present in X-ray reports. 

\noindent (2) \textit{TUH EEG Data} is an EEG dataset which provides variable length EEG recording and corresponding EEG report \cite{obeid2016temple} collected at Temple University Hospital. This dataset contains 16,950 sessions from 10,865 unique subjects. We preprocess the reports to extract the impression section of the report. We apply similar tokenization to these reports to extract tokens. We only keep the tokens with 3 or more occurrences.

\noindent (3) \textit{Massachusetts General Hospital (MGH) EEG Data} This is another EEG reports dataset which was used to evaluate our methods which was collected at large hospital in United States and contains EEG recordings paired with EEG reports written by clinicians. This dataset contains 12,980 deidentified EEG recordings paired with text reports.  We apply similar preprocessing steps to clean the reports from this dataset.

The data statistic are summarized in Table~\ref{tab:stats}. 

\begin{table}[h]
\centering
\caption{Dataset Statistics}
\label{tab:stats}
\resizebox{1\columnwidth}{!}{
\begin{tabular}{lccc}
\toprule
& IU X-Ray  & TUH EEG & MGH EEG \\ \hline
Number of Patients    &  3,996 &  10,890    &    10,865     \\ 
Number of Reports  &  7,470 &  12,980          &     16,950    \\ 
Total EEG length     & - &    4,523 hrs    &     3,452 hrs     \\ 
Total number of Final Tokens & 1235  &    2987    &     2675   \\ \hline
\end{tabular}}
\end{table}

\noindent\textbf{Baselines}: For IU X-ray image data, we compared \mname with these following baselines. We use DenseNet~\cite{huang2017densely} as the CNN model for extracting features for all variants of \mname models for fair comparison. 

\begin{enumerate}[leftmargin=*]
    \item \textbf{CNN-RNN}~\cite{vinyals2015show} passes the image through a CNN to obtain visual features and then passes to an LSTM to generate text reports.
    \item \textbf{Adaptive Attention}~\cite{lu2017knowing} uses adaptive attention  to produces context vectors and then generate text reports via LSTM .
    \item \textbf{HRGR}~\cite{li2018hybrid}  uses reinforcement learning to either generate a text report or retrieve a report from a template database.
    \item \textbf{KERP}~\cite{li2019knowledge} uses a graph transformer-based neural network to generate reports with a template database based approach.
    \item \textbf{AG}~\cite{jing2017automatic} first generates the tags associated with X-ray reports then generates reports based on those tags and visual features. 
\end{enumerate}

Likewise, for EEG datasets, we consider the following baselines.
\begin{enumerate}[leftmargin=*]
    \item \textbf{Mean-pooling(MP)} \cite{venugopalan2014translating} uses CNN to extract features for different EEG segments and then combine them using mean pooling. The output feature vectors are then passed to a 2-layer LSTM to generate text reports.
    \item \textbf{S2VT}  \cite{venugopalan2015sequence}  applies a seq-to-seq model which reads CNN outputs using an LSTM and then produce text with another LSTM.
    \item \textbf{Temporal Attention Network(TAM)} \cite{yao2015describing}  uses CNN to learn EEG features and then passes them to a decoder equipped with temporal attention which allows focusing on different EEG segments to produce the text report.
    \item \textbf{Soft Attention(SA)}~\cite{bahdanau2014neural} uses a soft attention mechanism to allow the decoder for focusing on EEG feature representations.
    \item \textbf{EEG2text}\cite{EEG2text} develops a template based approach to generate EEG reports using a hybrid model of CNN and LSTM.
\end{enumerate}

\noindent\textbf{Metrics:} 
To evaluate report generation quality, we use BLEU\cite{papineni2002bleu} and CIDEr \cite{vedantam2015cider} which are commonly used to evaluate language generation tasks. In addition, we also qualitatively evaluate the generated texts via a user study with doctors.

\noindent\textbf{Training Details}
For all models, we split the data into  train, validation, test set with 70\%, 10\%, 20\% ratio. There is no overlap between patients between train, validaation and test sets. The word embeddings which are used in the editing module were pre-trained specifically for each dataset. 

\noindent\textbf{Implementation Details}~\label{appendix:implementation}
We implemented \mname in PyTorch 1.2 \cite{Paszke2017-sg}.We use ADAM \cite{kingma2014adam} with batch size of 128 samples. We use a machine equipped with Intel Xeon e5-2640, 256GB RAM, eight Nvidia Titan-X GPU and CUDA 10.0. For ADAM to optimize all models and the learning rate is selected from [2e-3, 1e-3, 7.5e-4] and $\beta_1$ is selected from [0.5, 0.9].  We train all models for 1000 epochs. We start to half the learning rate every 2 epochs after epoch 50. We used 10\% of the dataset as a validation set for tuning hyper-parameters of each model. We searched for different model parameters using random search method. While preprocessing the text reports, if words were excluded, then a special "UNKNOWN" token is used to represent that word. Word embeddings were used with the seq2seq model in the editing module of \mname. Word embedding are typically used with such models to provide a fixed length vector to the LSTM model. We used pretrained word embeddings in our training procedure.\\

\noindent\textbf{Pretraining CNN for X-ray data}.
It has been shown that pretraining of neural networks leads to better classification performance in various tasks. In other image captioning tasks, often ResNets pretrained on imagenet dataset is used instead of retraining the entire network from scratch. So we also pretrained a DenseNet~\cite{huang2017densely} model with publicly available ChestX-ray8~\cite{wang2017chestx} dataset on multi-label classification.
ChestX-ray8 dataset consists of 108,948 frontal-view X-ray images of 32,717 unique patients with each image labeled with occurrence of 14 common thorax diseases where labels were text-mined from the associated radiological reports using natural language processing.\\

\noindent\textbf{Encoder CNN Details for EEG data}.
Usually the input EEG is 25-30minutes long, we divide EEG into 1 minute segments. This chunking operation leads [19x6000x30] dimension input for 30 minute length EEG where there are 19 channels and 6000 data points for time(100Hz, 60second).
Each of the 19x600 is passed through a CNN architecture which can accept multi-channel input. This CNN is composed multiple convolution, batch normalization, max-pooling blocks. The output of this CNN is 1x512 dimension feature vector which is obtained at last layer of the network which is a fully connected layer to obtain the final representation.

These are the steps of the operations for the CNN with EEG input. In the following notations, Conv2D refers to a 2D convolution operation. DepthwiseConv2D refers to depthwise spatial convolution. Separable Conv2D refers to separable convolutions consisting  of a depth wise spatial convolution followed by a pointwise convolution. The following operations describe the CNN for processing the EEG input. (1)Input EEG  -> (C,T) (2) Reshape -> (1,C,T) (3) Conv2D -> (F1, C, T), kernel size = 64, filter = 8 [here we denote C = number of channels, T = number of time points, F1= filter size](4) Batch Normalization (5) DepthwiseConv2D , number of spatial filters = 2 (6) Batch Normalization (7) Activation , ReLU (8) AveragePool2D, pool size=(1, 4) (9) Dropout, Dropout Rate = 0.5 (10) Separable Conv2D, filters =16 (11) Batch Normalization (12) Activation -> ReLU (13) AveragePool2D: \text{pool size = (1,8)} (14) Dropout: Dropout Rate = 0.5 (15) Dense.

\subsection{Anchor words used as classification labels}~\label{appendix:anchor_word}
Anchor words are the words which are used by our method \mname to trigger auto-completion by retrieving and editing the sentences to produce the final report. These words are critical because these are used to extract different candidate sentences. We have listed different anchor words which are used in our experiments. 
For image, we used labels in CheXpert  \cite{irvin2019chexpert} as the anchor words for the X-ray report completion. For EEG, we use a list of terms obtained from American Clinical Neurophysiology Society (ACNS)~\cite{hirsch2013american} guidelines. The two sets of keywords are listed below.

\begin{enumerate}[leftmargin=*]
\item \textbf{X-ray anchor words} include the following ones: No Finding; Enlarged Cardiomediastinum; Cardiomegaly; Lung Lesion; Airspace Opacity; Edema; Consolidation ; Pneumonia ; Atelectasis; Pneumothorax ; Pleural Effusion; Pleural Other; Fracture.
\item \textbf{EEG anchor words} include Normality, Sleep, Generalized Slowing, Focal Slowing, Epileptiform Discharges, Drowsiness, Spindles, Vertex Waves, Seizure. 
\end{enumerate}

\subsection{Results}

\noindent\textbf{(A). \mname can generate higher quality clinical reports}

We compare \mname with state-of-the-art baselines using the following experiments: 
\begin{enumerate}[leftmargin=*]
\item Report level auto-completion with predefined anchor words.
\item Report level auto-completion without predefined anchor words (i.e., anchor words are predicted). This experiment evaluates the scenario of fully automated report generation.
\item Sentence level auto-completion. This experiment simulates the real-world report auto-completion behavior where the recommendation is provided sentence by sentence.
\end{enumerate}
Table~\ref{tb:autocompletion} summarizes the \textbf{report level performance} on both X-ray image and EEG datasets. \mname (predicted anchor words) outperforms the best baselines with a 17-30\% improvement in CIDEr, which confirms the effectiveness of the retrieval from the prototype repository. We can also see with interactive guidance of anchor words from clinicians,  \mname (defined anchor words) provides an even better performance, which shows the importance of human input for report generation. To further understand the behavior of individual modules, we evaluate \mname without edit module (only sentence retrieval from existing reports), which still achieves better performance in CIDEr than baselines but is much lower than \mname (predicted anchor words) utilizing both retrieval and edit modules. 

With sentence-by-sentence interactive report auto-completion with anchor words and prefix text, the performance of \mname can be further improved.
We evaluated \mname with varying numbers of anchor words and prefix sentences to understand the effect of the increasing number of anchor words. We used 1-5 anchor words with \mname. We also used prefix sentences with variable length. We present these sentence-level auto-completion results in table \ref{tab:perf_sentence_Impression}. As the results show the with increasing the number of anchor words, we can obtain higher scores. We observe that with increasing the number of anchor words the performance of \mname increases 1-2\%. In real-world deployed version of \mname, clinicians can provide more input(anchor words) to the system to obtain more accurate results which is an advantage over current baselines where clinicians do not have control over the report generation.

\begin{figure*}[h!]
\centering
    \includegraphics[width=0.9\textwidth]{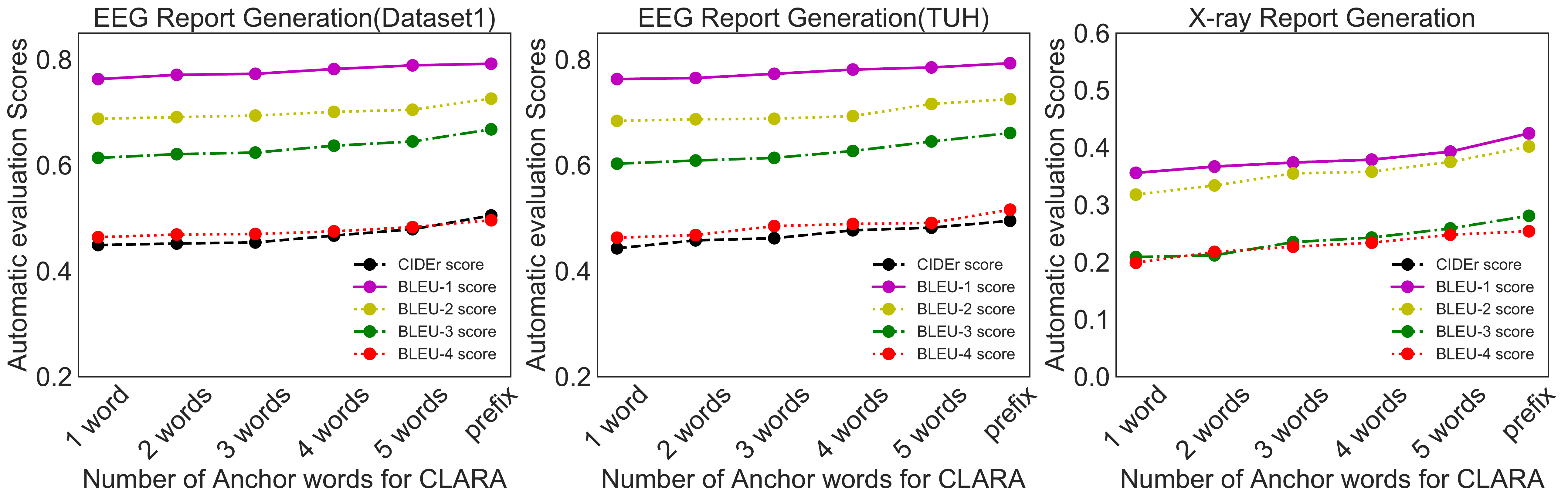}
    \caption{This plot shows that the increasing CIDEr and BLEU scores for EEG reports and X-ray report generation with an increasing number of anchor words. This increasing trend of the scores with an increasing number of anchor words indicates that anchor words help guide \mname to produce higher-quality reports. As clinicians provide more anchor words, \mname can extract better candidate sentences and edit the sentences}
    \label{fig:eeg_example_2}
\end{figure*}

\begin{table}[h!]
\centering
  \caption{Sentence level completion  for EEGs and X-rays}
  \label{tab:perf_sentence_Impression}
    \resizebox{1\columnwidth}{!}{
  \begin{tabular}{c|lccccc}
    \toprule
    Dataset & Method & CIDEr & BLEU-1 & BLEU-2  & BLEU-3 & BLEU-4 \\
      \midrule
    \multirow{9}{*}{IU X-ray (image)} &\multirow{1}{*}{CNN-RNN} &   0.294 & 0.216 & 0.124 & 0.087 & 0.066\\ 
    &\multirow{1}{*}{Adaptive Attention}  &  0.295 &  0.220 & 0.127 & 0.089 & 0.068 \\
    &\multirow{1}{*}{AG}~\cite{jing2017automatic} &  0.277 & 0.455 & 0.288 & 0.205 & 0.154\\
    &\multirow{1}{*}{HRGR}~\cite{li2018hybrid}  & 0.343	& 0.438 & 0.298 &  0.208 & 0.151 \\
                       
  &\multirow{1}{*}{\mname} (1 anchor word)  &  0.356 &  0.471 &  0.318 &  0.209 &  0.199 \\
  &\multirow{1}{*}{\mname} (2 anchor words) &  0.367 &  0.484 &  0.334 &  0.212 &  0.218 \\
  &\multirow{1}{*}{\mname} (3 anchor words) &  0.374 & 0.488 &  0.355 &  0.235 & 0.227 \\
  &\multirow{1}{*}{\mname} (4 anchor words)  &  0.379 &  0.495 &  0.358 &  0.243 &  0.234 \\
    &\multirow{1}{*}{\mname} (5 anchor words)  &  0.393 &  0.498 &  0.375 & 0.259 & 0.248 \\
    &  \multirow{1}{*}{\mname}(with prefix) &  0.425 &  0.512 &  0.402 &  0.281 & 0.254 \\
    \midrule\midrule
    \multirow{9}{*}{MGH (EEG)} & \multirow{1}{*}{MP} \cite{venugopalan2014translating} &  0.371 & 0.715 & 0.634 & 0.561 & 0.448\\ 
    & \multirow{1}{*}{S2VT} \cite{venugopalan2015sequence}  &  0.321 & 0.748 & 0.623 & 0.531 & 0.469 \\  
    & \multirow{1}{*}{TAM} \cite{yao2015describing} &  0.345 & 0.748 & 0.672 & 0.593 & 0.381 \\ 
    &  \multirow{1}{*}{SA} \cite{bahdanau2014neural} &  0.353 & 0.689 & 0.634 & 0.573 &  0.484 \\
    &  \multirow{1}{*}{EEG2text} \cite{EEG2text} &  0.386 & 0.731 & 0.719 & 0.562 &  0.453 \\ 
    &  \multirow{1}{*}{\mname}(1 anchor word) &  0.443 &  0.763 &  0.684 &  0.603 & 0.463 \\
    &  \multirow{1}{*}{\mname}(2 anchor words) & 0.458 &  0.765 &  0.687 &  0.609 & 0.468 \\
    &  \multirow{1}{*}{\mname}(3 anchor words) &  0.462 &  0.773 &  0.688 &  0.614 & 0.485 \\
    &  \multirow{1}{*}{\mname}(4 anchor words) &  0.477 &  0.781 &  0.693 &  0.627 & 0.489 \\
    &  \multirow{1}{*}{\mname}(5 anchor words) &  0.482 &  0.785 &  0.716 &  0.645 & 0.491 \\
    &  \multirow{1}{*}{\mname}(with prefix) &  0.495 &  0.793 &  0.725 &  0.661 & 0.516 \\
    \midrule  
    \midrule 
 
    \multirow{9}{*}{TUH (EEG)}& \multirow{1}{*}{MP\cite{venugopalan2014translating}} & 0.368 & 0.643 & 0.579 & 0.462 & 0.364 \\    
    & \multirow{1}{*}{S2VT \cite{venugopalan2015sequence}} &  0.371 & 0.725 & 0.634 & 0.545 & 0.441 \\\
    &  \multirow{1}{*}{TAM \cite{yao2015describing}}  &  0.385 & 0.719 & 0.646 & 0.503 & 0.469\\
    &   \multirow{1}{*}{SA  \cite{bahdanau2014neural}}  &  0.353 &  0.738 & 0.621 & 0.524 & 0.432 \\
    &  \multirow{1}{*}{EEG2text} \cite{EEG2text} &  0.368 & 0.723 & 0.678 & 0.609 &  0.457 \\ 
    &  \multirow{1}{*}{\mname}(1 anchor words) &  0.449 &  0.763 &  0.688 &  0.614 & 0.464 \\
    &  \multirow{1}{*}{\mname}(2 anchor words) &  0.452 &  0.771 &  0.691 &  0.621 & 0.469 \\
    &  \multirow{1}{*}{\mname}(3 anchor words) &  0.454 &  0.773 &  0.694 &  0.624 & 0.470 \\
    &  \multirow{1}{*}{\mname}(4 anchor words) &  0.467 &  0.782 &  0.701 &  0.637 & 0.475 \\
    &  \multirow{1}{*}{\mname}(5 anchor words) &  0.479 &  0.789 &  0.705 &  0.645 & 0.483 \\
    &  \multirow{1}{*}{\mname}(with prefix) &  0.505 &  0.792 &  0.726 &  0.668 & 0.496 \\
    \bottomrule
  \end{tabular}}
  \vskip -1em
\end{table}

\begin{table}[h!]
\centering
  \caption{{Accuracy of disease phenotype prediction based on generated reports}}
  \label{tab:phenotype_classification}
  \resizebox{1\columnwidth}{!}{
  \begin{tabular}{c|lcc}
    \toprule
    Dataset & Method
    &  Averaged Accuracy & PR-AUC \\
    \midrule
    \multirow{4}{*}{IU X-ray (image)}& \multirow{1}{*}{CNN-RNN} &   0.804 & 0.709 \\ 
    &\multirow{1}{*}{Adaptive Attention}  &  0.823  & 0.723 \\
    &\multirow{1}{*}{\mname} (predicted anchor words)  & \bf 0.871  & \bf 0.796 \\
    &\multirow{1}{*}{\mname} (defined anchor words) & \bf 0.894 & \bf 0.804 \\
    \midrule\midrule
    \multirow{6}{*}{MGH} & \multirow{1}{*}{MP} \cite{venugopalan2014translating} &  0.745 & 0.724\\ 
    & \multirow{1}{*}{S2VT} \cite{venugopalan2015sequence}  &  0.773  & 0.738\\  
    & \multirow{1}{*}{TAM} \cite{yao2015describing} &  0.761  & 0.713 \\ 
    &  \multirow{1}{*}{SA} \cite{bahdanau2014neural} &  0.743  & 0.716 \\
    &  \multirow{1}{*}{EEG2text} \cite{EEG2text} &   0.784  & 0.748 \\
    &  \multirow{1}{*}{\mname}(predicted anchor words) & \bf 0.835 & \bf 0.803 \\
    &  \multirow{1}{*}{\mname}(defined anchor words)&  \bf 0.861 & \bf 0.814 \\
    \midrule  
    \midrule 
   \multirow{6}{*}{TUH} & \multirow{1}{*}{MP  \cite{venugopalan2015sequence}} & 0.758 & 0.697     \\               
    & \multirow{1}{*}{S2VT \cite{venugopalan2015sequence}} &  0.764 & 0.683 \\
    &  \multirow{1}{*}{TAM} \cite{yao2015describing} &  0.782 & 0.698\\
    &   \multirow{1}{*}{SA} \cite{bahdanau2014neural}  &  0.781 & 0.701 \\
    &  \multirow{1}{*}{EEG2text} \cite{EEG2text} &   0.793  & 0.735 \\
    &  \multirow{1}{*}{\mname}(predicted anchor words) & \bf 0.827  \bf & 0.786 \\
    &  \multirow{1}{*}{\mname}(defined anchor words) & \bf 0.834  & \bf 0.804 \\

    \bottomrule
  \end{tabular}}
  \vspace{-0.20cm}
\end{table}


\begin{figure*}[h!]
\centering
    \includegraphics[width=0.9\textwidth]{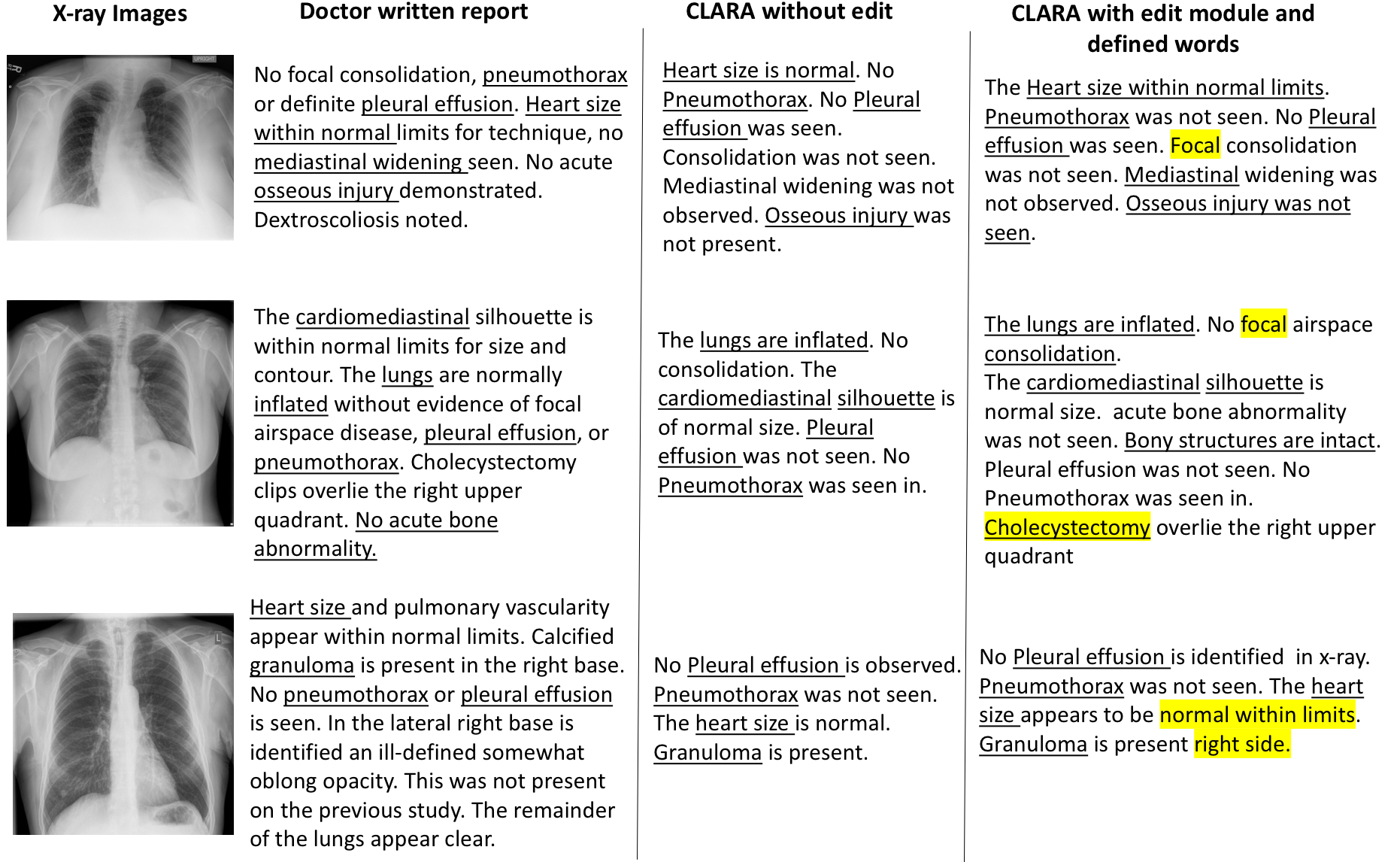}
    \vskip -1em
    \caption{Examples of X-ray reports (doctor written report vs. generated reports). The second column shows the doctor written reports. The third and fourth columns show the reports generated by \mname without edit module (just retrieval of most relevant sentences) and \mname with edit module using defined anchor words (our best model), respectively. Underlined text indicates matched terms of the generated text and ground truth reports. The highlighted words show the changes added by the edit module of \mname. These generated reports show that our method \mname is able to write reports that are similar to the doctor's written report. \mname without the edit module is also capable of extracting good candidate sentences.}
    \vskip -1em
    \label{fig:example_1}
\end{figure*}

\begin{figure*}[h!]
\centering
    \includegraphics[width=0.9\textwidth]{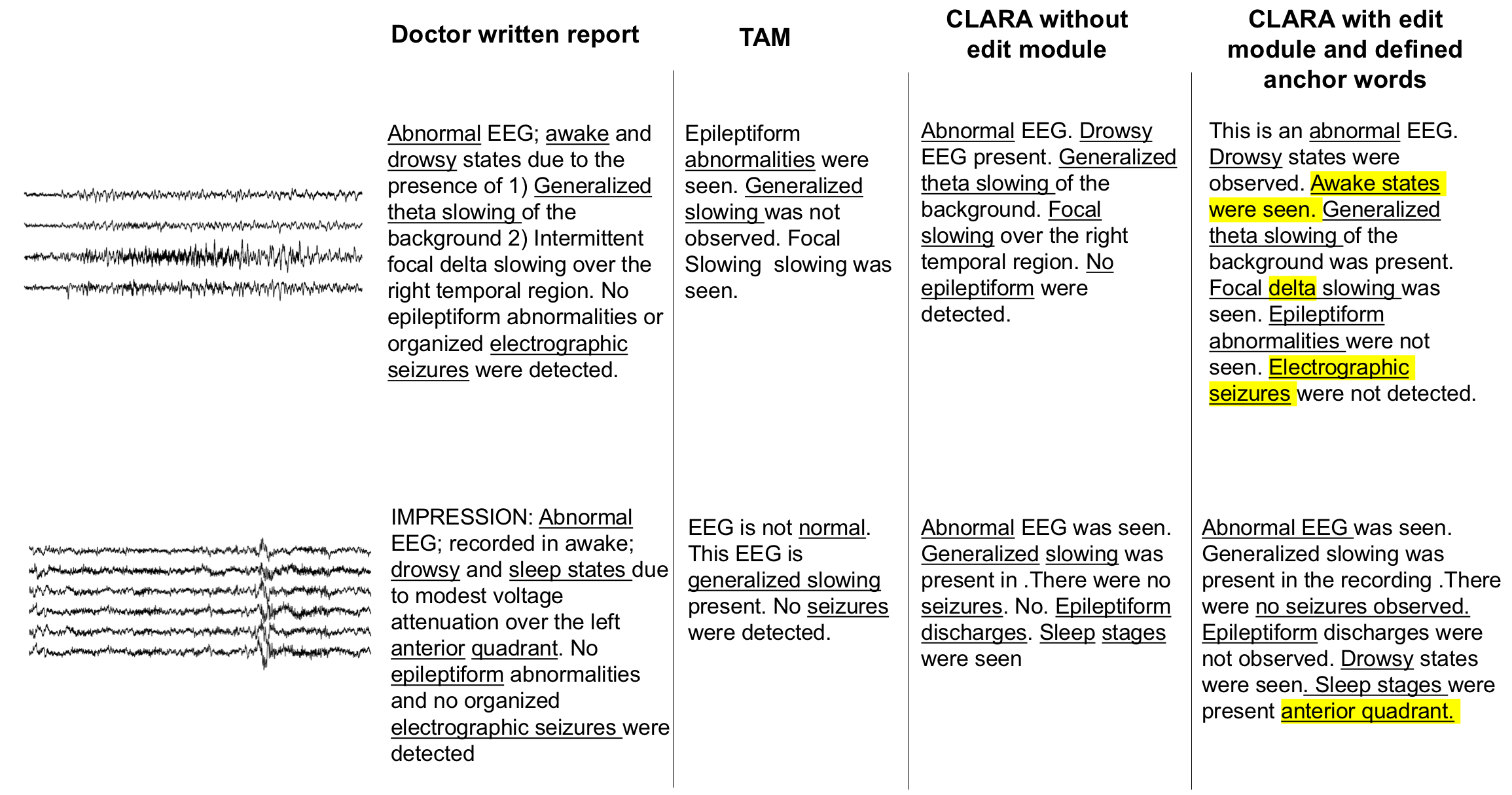}
    \vskip -1em
    \caption{An example of EEG report generation. The first column shows the ground truth text report. The second column shows the report generated by TAM \cite{yao2015describing} method. The third column shows the report generated \mname without edit module and the fourth column shows the report generated by \mname with edit module using defined anchor words. Underlined text indicates the correspondence of the generated text and ground truth reports. The highlighted words show the edit module of \mname adding more information to the retrieved report. Noticeably CLARA with edit module can capture specific information such as ``Delta slowing'', ``anterior quadrant''. These changes introduced by the edit module showcase the importance of retrieve and edit module for producing a module}
    \vskip -1em
    \label{fig:eeg_example_2}
\end{figure*}

\begin{table}[h!]
\centering
  \caption{Report level completion tasks on image data (IU X-ray)  and EEG time series data (MGH and TUH) for Impression Section generation task }~\label{tb:autocompletion}
  \resizebox{1\columnwidth}{!}{
  \begin{tabular}{c|lccccc}
    \toprule
    Dataset & Method & CIDEr & BLEU-1 & BLEU-2  & BLEU-3 & BLEU-4 \\
    \midrule
    \multirow{6}{*}{IU X-ray (image)} & \multirow{1}{*}{CNN-RNN} &   0.294 & 0.216 & 0.124 & 0.087 & 0.066\\ 
    &\multirow{1}{*}{Adaptive Attention}  &  0.295 &  0.220 & 0.127 & 0.089 & 0.068 \\
    &\multirow{1}{*}{AG}~\cite{jing2017automatic} &  0.277 & 0.455 & 0.288 & 0.205 & 0.154\\
    &\multirow{1}{*}{HRGR}~\cite{li2018hybrid}  & 0.343	& 0.438 & 0.298 &  0.208 & 0.151 \\
    &\multirow{1}{*}{KERP} ~\cite{li2019knowledge}  & 0.280	& 0.482 & 0.325 &  0.226 & 0.162 \\
    &\multirow{1}{*}{\mname}(without edit module) &  0.317 &  0.421 &  0.288 &  0.201 &  0.142 \\
    &\multirow{1}{*}{\mname}(predicted anchor words)  & \bf 0.359 & \bf 0.471 & \bf 0.324 & \bf 0.214 & \bf 0.199 \\
    &\multirow{1}{*}{\mname}(defined anchor words) & \bf 0.374 & \bf 0.489 & \bf 0.356 & \bf 0.225 & \bf 0.234 \\
      \midrule\midrule
    
    \multirow{6}{*}{MGH  (EEG)} & \multirow{1}{*}{MP} \cite{venugopalan2014translating} &   0.367 & 0.714 & 0.644 & 0.563 & 0.443\\ 
    & \multirow{1}{*}{S2VT} \cite{venugopalan2015sequence}  &  0.319 & 0.741 & 0.628 & 0.529 &  0.462 \\  
    & \multirow{1}{*}{TAM} \cite{yao2015describing} &  0.334 & \bf 0.749 & 0.668 & 0.581 & 0.378\\ 
    &  \multirow{1}{*}{SA} \cite{bahdanau2014neural} &  0.348 & 0.684 & 0.629 & 0.568 & \bf 0.472 \\
    &  \multirow{1}{*}{EEG2text} \cite{EEG2text} &  0.372 & 0.742 & 0.728 & 0.587 &  0.381 \\ 

    &  \multirow{1}{*}{\mname}(without edit module) &  0.382 &  0.691 &  0.651 &  0.564 &  0.405 \\
    &  \multirow{1}{*}{\mname}(predicted anchor words) & \bf 0.419 &  0.742 & \bf 0.674 & \bf 0.594 & 0.452 \\
    &  \multirow{1}{*}{\mname}(defined anchor words) & \bf 0.443 & \bf 0.762 & \bf 0.684 & \bf 0.614 & \bf 0.464 \\
    \midrule  
    \midrule 
 
    \multirow{6 }{*}{TUH (EEG )}& \multirow{1}{*}{MP} \cite{venugopalan2014translating} & 0.363 & 0.645 & 0.578 & 0.459 & 0.361 \\    
                        
    & \multirow{1}{*}{S2VT}  \cite{venugopalan2015sequence}  &  0.364 & 0.724 & 0.613 & 0.543 & 0.438 \\
                          
    &  \multirow{1}{*}{TAM} \cite{yao2015describing}  &  0.384 & 0.714 & 0.647 & 0.492 & 0.461\\
                         
    &  \multirow{1}{*}{SA}  \cite{bahdanau2014neural}  &  0.341 &  0.736 & 0.619 & 0.519 & 0.420 \\
    &  \multirow{1}{*}{EEG2text} \cite{EEG2text} &  0.381 & 0.752 & 0.618 & 0.593 &  0.428 \\ 
    &  \multirow{1}{*}{\mname}(without edit module) &  0.368 &  0.725 &  0.634 &  0.573 &  0.423 \\
                       
    &  \multirow{1}{*}{\mname}(predicted anchor words) & \bf 0.399 & \bf 0.769 & \bf 0.635 & \bf 0.601 & \bf 0.455\\
    &  \multirow{1}{*}{\mname}(defined anchor words) & \bf 0.425 & \bf 0.784 & \bf 0.659 & \bf 0.624 & \bf 0.483\\

    \bottomrule
  \end{tabular}}
\end{table}

\noindent\textbf{(B). \mname provides accurate disease phenotyping}

We also evaluate the effectiveness of \mname in disease phenotype prediction. 
In particular, we train a character CNN  classifier\cite{zhang2015character} on original reports written by doctors to predict disease phenotypes.  This classifier is used to score the generated reports produced by different baselines and \mname. We measure the accuracy for predicting different disease phenotypes. 
\mname consistently outperforms baselines. 
This results for X-ray is in Table~\ref{tab:phenotype_classification}.




\noindent\textbf{(C). Clinical Expert Evaluation of \mname}
The results of our models were evaluated by an expert neurologist in terms of its usefulness for clinical practice. In this setup, we measured the quality score metric for the generated reports. We only evaluated the EEG report generation task in this experimental setting. We provided the experts with samples with doctor written reports, reports generated by best-performing baselines and \mname presented side by side. Clinicians were asked to provide a quality score in the range of 0-5. As shown in Figure~\ref{fig:userstudy}, \mname obtained an average quality score of 3.74 compared to TAM (best performing baseline)  obtaining an average quality score of 2.52. These results indicate that the reports produced by \mname are of higher clinical quality.
\begin{figure}[htb]
\centering
    \includegraphics[width=0.6\columnwidth]{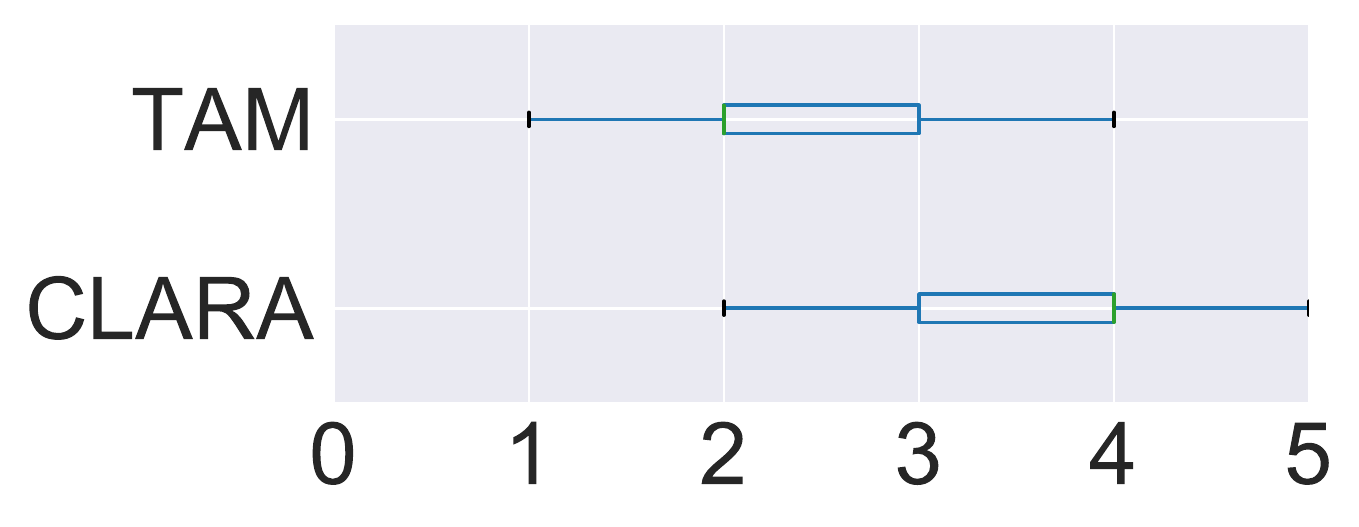}
    \vskip -1em
    \caption{Clinical expert assessment of generated reports. }
    \vskip -1em
    \label{fig:userstudy}
\end{figure}

\noindent\textbf{(D). Qualitative Analysis}

We show sample results of clinical report generation  using \mname in Figure \ref{fig:example_1}. Reports generated by \mname show significant clinical accuracy and granular understanding of the input image. As clinicians use \mname with different anchor words to generate the report, it ensures inclusion of important clinical findings such as ``Pleural effusion'', ``Pneumothorax''. Since anchor words are based on clinically significant findings, it enforces the report generation module to be clinically accurate. The third and fourth columns of the figure show the difference and changes introduced by the edit module. The edit module is  able to modify the retrieved sentences with more details. For example, ``focal consolidation'' in row 1, ``Chloecystectomy'' in row 2, ``Granuloma in right side'' are important edits performed by the edit module of \mname.

\section{Conclusion}
Medical report writing is important but labor-intensive for human doctors. Most existing works on medication generation focus on generating full reports without close human guidance, which is error-prone and does not follow clinical workflow. In this work, we propose \mname, a computational method for supporting clinical report auto-completion task, which interactively facilitates doctors to write clinical reports in a sentence by sentence fashion. 
At the core, \mname combines the information retrieval engine and neural networks to enable a powerful mechanism to retrieve most relevant sentences via retrieval systems then modify that using neural networks. Our experiments show that \mname can produce higher quality and clinically accurate reports. \mname outperforms a variety of compelling baseline methods across tasks and datasets with up to 35\% improvement in CIDEr and BLEU-4 over the best baseline. 

\section{Acknowledgments}
This work is part supported by National Science Foundation award IIS-1418511, CCF-1533768 and IIS-1838042, the National Institute of Health award NIH R01 1R01NS107291-01 and R56HL138415.
\section{Conclusion}
In this work, we proposed \mname, a doctor representation learning based on both patient representations from longitudinal patient EHR data and trial embedding from the multimodal trial description.  \mname leverages a dynamic memory network where the representations of patients seen by the doctor are stored as memory while trial embedding serves as queries for retrieving the memory. Evaluated on real world patient and trial data, we demonstrated via
trial enrollment prediction tasks that \mname can learn accurate doctor embeddings and greatly outperform state-of-the-art baselines. We also show by additional experiments that the \mname embedding can also be transferred to benefit the data insufficient setting (e.g., model transfer to less populated/newly explored country or from common disease to rare disease) that is highly valuable yet extremely challenging for clinical trials.

\section*{Acknowledgement}
This work was in part supported by the National Science Foundation award IIS-1418511, CCF-1533768 and IIS-1838042, the National Institute of Health award NIH R01 1R01NS107291-01 and R56HL138415.


\bibliographystyle{aaai}

\bibliography{sample}

\clearpage
\appendix
\section{Supplemetary}

\subsection{Lucene Details}~\label{appendix:lucene}
Lucene implements a variant of the Tf-Idf scoring model
\begin{itemize}[leftmargin=*]
    \item tf = term frequency in document = measure of how often a term appears in the document
    \item idf = inverse document frequency = measure of how often the term appears across the index
    \item coord = number of terms in the query that were found in the document
    \item lengthNorm = measure of the importance of a term according to the total number of terms in the field
    \item queryNorm = normalization factor so that queries can be compared
    \item boost (index) = boost of the field at index-time
    \item boost (query) = boost of the field at query-time
\end{itemize}

\begin{multline}
     score(q,d) = coord(q,d)  queryNorm(q) \\ \sum_{t \in q}
    ( (tf(t in d) idf(t)^2  t.getBoost()  norm(t,d) ) 
\end{multline}

\textbf{Factor	Description}:
\begin{enumerate}[leftmargin=*]
    \item tf(t ind):	Term frequency factor for the term (t) in the document (d).
    \item idf(t):	Inverse document frequency of the term.
    \item coord(q,d):	Score factor based on how many of the query terms are found in the specified document.
    \item queryNorm(q):	Normalizing factor used to make scores between queries comparable.
    \item t.getBoost():	Field boost.
    \item norm(t,d):	Encapsulates a few (indexing time) boost and length factors.
\end{enumerate}

\noindent\textbf{Lucene steps from query to output}
In this section, we describe some details of the search engine behind Lucene. Usually, query is passed to the Searcher of the Lucene engine, beginning the scoring process. Then the Searcher uses Collector for the scoring and sorting of the search results. These important objects are involved in a search: (1) The Weight object of the Query: this is an internal representation of the Query that allows the Query to be reused by the Searcher. (2) The Searcher that initiated the call.(3) Filter for limiting the result set. (4) Sort object for specifying the sorting criteria for the results when the standard score based sort method is not desired.

\noindent\textbf{Simulated auto-completion}:
The auto-completion system requires a trigger from the user to initiate the process. These triggers are initialized with prefix or anchor words provided by the user. In real-world a deployed version of our model, doctors will provide anchor words ore prefix to \mname to trigger completion of the sentences. But while developing the method, we can not expect to train the model with input from clinicians. So we created a simulated environment where the anchor words are predefined for each report. These anchor word creation steps are detailed in the above section.

\end{document}